\documentclass[letterpaper,twocolumn]{article}
\usepackage[T1]{fontenc}
\usepackage{newtxtext,newtxmath}
\usepackage{helvet}
\usepackage{courier}
\usepackage{xcolor}
\usepackage[hyphens]{url}
\usepackage{graphicx}
\usepackage{booktabs}
\usepackage{amsmath}
\usepackage{newtxmath}
\usepackage{natbib}
\usepackage{caption}
\usepackage{placeins}
\usepackage[margin=1in]{geometry}
\newcommand{\equalcontrib}{\textsuperscript{*}}
\newcommand{\corresponding}{\textsuperscript{\dag}}

\newcommand{\ww}[1]{\makebox[0.96cm][c]{#1}}

\title{BioVLN: A Simulation Platform for Visual Language Navigation in Biomedical Laboratories}

\author{
    Zhe~Liu\equalcontrib\textsuperscript{1,2},
    Quan~Lu\equalcontrib\textsuperscript{1,2},
    Zhaohui~Du\equalcontrib\textsuperscript{1,2},
    Zhe~Wang\corresponding\textsuperscript{1,2},
    Huanbo~Jin\textsuperscript{1,2},\\
    Jiaming~Gu\textsuperscript{1,2},
    Qi~Wang\textsuperscript{3},
    Ting~Xiao\textsuperscript{3},
    Minting~Pan\textsuperscript{4},
    Dongzhan~Zhou\textsuperscript{4}
}

\date{
    \vspace{-4mm}
    \textsuperscript{1}Key Laboratory of Smart Manufacturing in Energy Chemical Process, MoE,
    East China University of Science and Technology, Shanghai, China\\
    \textsuperscript{2}Department of Computer Science and Engineering,
    East China University of Science and Technology, Shanghai, China\\
    \textsuperscript{3}Department of Laboratory Medicine, Ruijin Hospital,
    Shanghai Jiao Tong University School of Medicine, Shanghai, China\\
    \textsuperscript{4}AI for Science Center, Shanghai AI Laboratory, Shanghai, CN\\[4pt]
    \texttt{wangzhe@ecust.edu.cn}\\[2pt]
    \texttt{https://github.com/ActiveButterflies/BioVLN}
}
\begin{document}
\maketitle
\begingroup
\renewcommand{\thefootnote}{}
\footnotetext{*\hspace{0.2em}Equal contribution. \textsuperscript{\dag}\hspace{0.2em}Corresponding author.}
\endgroup

\begin{abstract}
Biomedical laboratory robots must navigate to instruments before performing experimental procedures. Existing embodied navigation platforms are designed for household environments and treat a target as an object center or an arbitrary nearby position. This representation is inadequate for laboratory instruments, which must be approached from their operating side while maintaining safe clearance from surrounding equipment. We introduce BioVLN, a simulation platform for developing and evaluating visual-language navigation agents in biomedical laboratories. BioVLN represents each instrument with three regions: its physical body, a surrounding clearance region, and an operation area in front of the usable side. This model is applied consistently to scene generation, target placement, navigation evaluation, and safety analysis, so success depends on reaching a position from which the instrument can be accessed. BioVLN supports procedural scene generation and manually designed environments, producing 47 scenes and 1{,}667 episodes. Standardized navigation and reinforcement-learning interfaces enable trajectory collection and policy training. Experiments show that geometric exploration reaches 74.4--87.5\% success, while sampling multiple valid positions in the operation area improves success to 83.3--92.5\% and reduces unsafe proximity.
\end{abstract}

\section{Introduction}

Language-conditioned embodied navigation, including visual-language navigation (VLN) and ObjectNav, has progressed rapidly through standardized simulators, large-scale indoor datasets, and increasingly capable semantic navigation agents. Habitat and AI2-THOR established reproducible environments for embodied learning, while ProcTHOR enabled large-scale procedural scene generation~\cite{savva2019habitat,kolve2017ai2,deitke2022procthor}. ObjectNav has similarly evolved from geometric and learned exploration to open-vocabulary methods that use vision-language models and scene-level reasoning to locate unseen targets~\cite{batra2020objectnav,yitzhak2022cows,yokoyama2024vlfm,yin2024sg}. These platforms and benchmarks are primarily designed for homes and offices, where success is defined by reaching the target object or a nearby position.

Biomedical laboratories require a more functional definition of navigation success. Laboratory robots must approach instruments before manipulation or experimental execution can begin~\cite{holland2020automation,burger2020mobile}. A centrifuge must be reached from its control side, a refrigerator from its door, and a benchtop instrument with sufficient clearance from adjacent equipment. The destination is not merely the instrument location, but a position from which it can be safely accessed. This requirement is further complicated by dense, workflow-dependent layouts and specialized instruments that can be difficult for general-purpose vision-language models to recognize.

Current navigation formulations do not explicitly capture these constraints. Point-based or proximity-based goals may indicate that an object has been reached, but they do not encode its operating direction or usable approach region. Standard metrics measure goal reaching and path efficiency without quantifying clearance from surrounding equipment. Mainstream scene-generation pipelines also lack a unified representation of instrument-specific operating directions, approach regions, and safety constraints. These limitations leave a gap between semantic object search and navigation that can support downstream laboratory interaction.

We introduce BioVLN, a simulation platform that formulates biomedical laboratory navigation around operational accessibility. Its central abstraction is a three-zone operational envelope comprising the instrument body, a surrounding clearance region, and an operation area in front of the usable side. This abstraction is applied consistently to scene construction, goal placement, episode generation, navigation evaluation, and trajectory-level safety analysis.

Our contributions are summarized as follows:

\begin{itemize}
\item We present BioVLN, an extensible simulation platform for visual-language navigation in biomedical laboratories. It introduces operational-face goals and a three-zone spatial model that couples instrument geometry, approach accessibility, and safety constraints throughout scene generation and evaluation.

\item We develop a dual-entry scene pipeline supporting both procedural construction and designer-authored scene import through the Blender-based LabScene Annotation Toolkit (LSAT). The platform provides 47 scenes and 1{,}667 episodes, with standardized trajectory-recording and Gym-compatible interfaces for zero-shot evaluation and policy learning.

\item We establish a six-method benchmark across three laboratory datasets using both navigation and safety metrics. Results show that sampling multiple valid positions within the operation area improves success to 83.3--92.5\% while reducing unsafe proximity, and reveal that vision-language navigation is strongly affected by the visual quality of rendered laboratory assets.
\end{itemize}

\begin{figure*}[t]
\centering
\includegraphics[width=\textwidth]{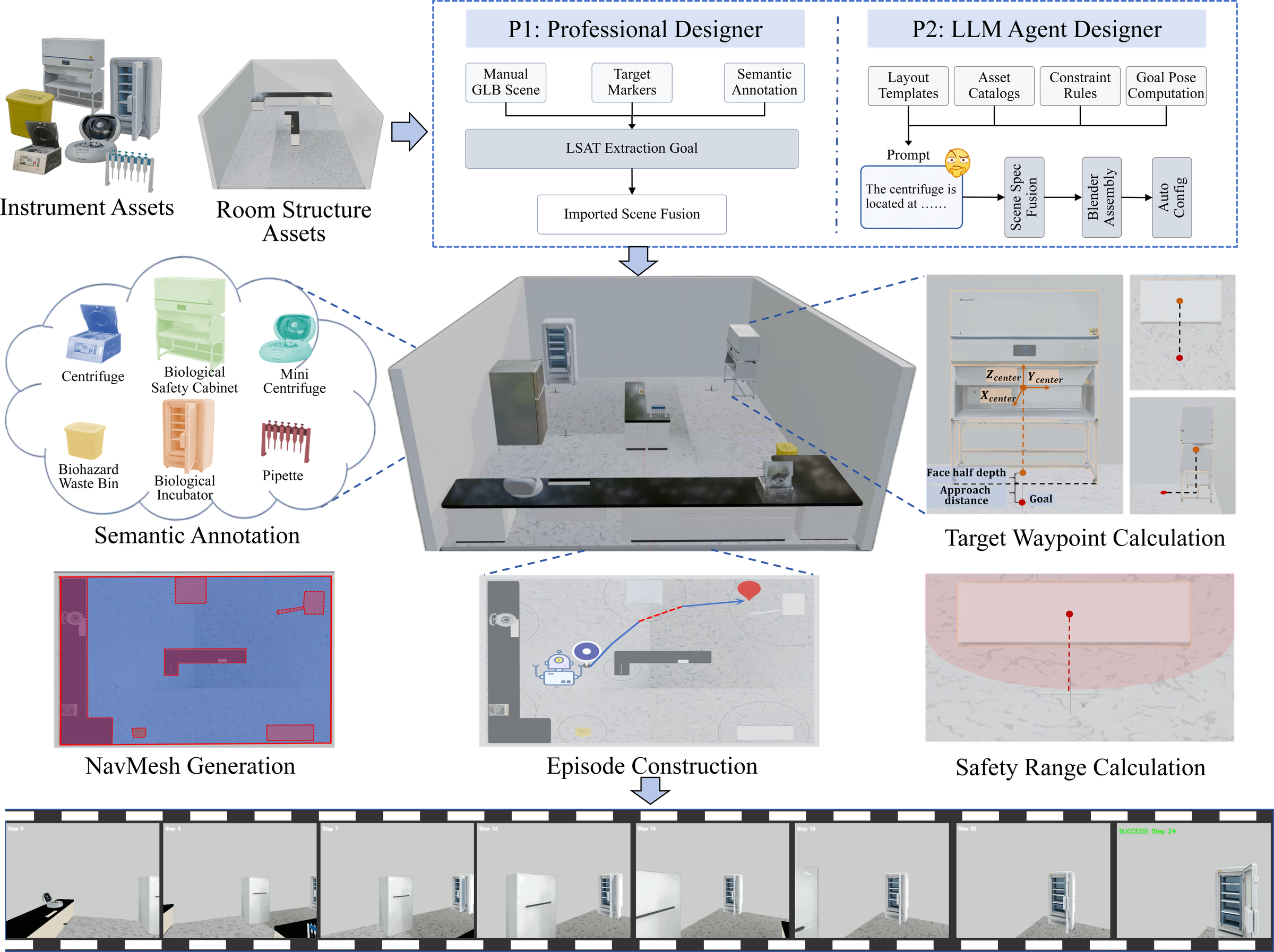}
\caption{The BioVLN platform architecture. P1 (left): designer-authored scene import, where manually built GLB scenes annotated via LSAT are processed. P2 (right): procedural generation, where instrument catalogs and layout templates drive automated construction. Both paths converge into a shared benchmark pipeline with six-method evaluation and safety metrics (MCR, VRT).}
\label{fig:pipeline}
\end{figure*}

\section{Related Work}

\subsection{Embodied Navigation Platforms}

The embodied AI community has developed several high-fidelity simulation platforms, evolving from fixed scene sets (Habitat-Sim~\cite{savva2019habitat}, AI2-THOR~\cite{kolve2017ai2}, Gibson~\cite{xia2018gibson}) to real-world building scans (Matterport3D~\cite{chang2017matterport3d}, HM3D~\cite{ramakrishnan2021habitat}), interactive manipulation (iGibson~\cite{li2021igibson}, BEHAVIOR~\cite{srivastava2022behavior}), large-scale procedural generation (ProcTHOR~\cite{deitke2022procthor}), and sim-to-real transfer (RoboTHOR~\cite{deitke2020robothor}). All existing platforms target residential or office domains and provide no mechanisms for evaluating operational safety or instrument-specific spatial constraints. Table~\ref{tab:platform_overview} contrasts BioVLN with representative platforms.

\begin{table*}[t]
\centering
\caption{Overview of existing embodied navigation platforms ($\checkmark$ = supported, $\times$ = not supported)}
\label{tab:platform_overview}
{\small
\setlength{\tabcolsep}{2pt}
\begin{tabular*}{\textwidth}{@{\extracolsep{\fill}}c c c c c c@{}}
\toprule
& Habitat & AI2-THOR & ProcTHOR & iGibson & BioVLN \\
& \cite{savva2019habitat} & \cite{kolve2017ai2} & \cite{deitke2022procthor} & \cite{li2021igibson} & (ours) \\
\addlinespace
\midrule
Domain & Home/Office & Home & Home & Home & \textbf{Laboratory} \\
Procedural scene generation      & $\times$ & $\times$ & $\checkmark$ & $\times$ & $\checkmark$ \\
Designer-authored scene import   & $\times$ & $\times$ & $\times$ & $\times$ & $\checkmark$ \\
LLM-assisted layout design       & $\times$ & $\times$ & $\times$ & $\times$ & $\checkmark$ \\
Operational-face goal definition & $\times$ & $\times$ & $\times$ & $\times$ & $\checkmark$ \\
Operational accessibility model  & $\times$ & $\times$ & $\times$ & $\times$ & $\checkmark$ \\
Safety metrics (MCR, VRT)        & $\times$ & $\times$ & $\times$ & $\times$ & $\checkmark$ \\
Blender-based scene annotation   & $\times$ & $\times$ & $\times$ & $\times$ & $\checkmark$ \\
\bottomrule
\end{tabular*}}
\end{table*}

\subsection{Object-Goal Navigation}

Object-goal navigation (ObjectNav)~\cite{batra2020objectnav} tasks an agent with navigating to any instance of a specified object category. The field has progressed from classical geometry-based exploration to learned policies (PONI~\cite{ramakrishnan2022poni}), semantic exploration (SemExp~\cite{singh2020object}), and, most recently, zero-shot VLM-guided approaches (ZSON~\cite{majumdar2022zson}, CoW~\cite{yitzhak2022cows}, SG-Nav~\cite{yin2024sg}). A fundamental limitation persists across all formulations: goals remain point targets at object centroids without orientation or affordance constraints. For laboratory instruments, the operational face (e.g., a refrigerator's front door versus its rear panel) determines whether a navigation trajectory is functionally useful. BioVLN reformulates the goal as an operational approach position with explicit affordance-facing direction.

\subsection{Frontier-Based Exploration}

Frontier-based exploration, introduced by Yamauchi~\cite{yamauchi1997frontier}, navigates toward boundaries between explored free space and unknown regions. The paradigm has evolved from hand-crafted heuristics to learned neural variants (Active Neural SLAM~\cite{chaplot2020learning}, Neural Topological SLAM~\cite{chaplot2020neural}) and large-scale RL (DD-PPO~\cite{wijmans2020ddppo}). Learned methods require environment-specific training and do not transfer zero-shot to domains with distinct visual and geometric statistics. We adopt a purely geometric frontier method as a training-free baseline.

\subsection{Vision-Language Models for Navigation}

Leveraging large vision-language models (VLMs) for zero-shot semantic navigation has attracted considerable recent interest. VLFM~\cite{yokoyama2024vlfm} scores frontier views with BLIP-2~\cite{li2023blip} and GPT-4o~\cite{hurst2024gpt} for semantic value map construction. InstructNav~\cite{long2024instructnav} leverages LLMs for instruction-following zero-shot navigation. SG-Nav~\cite{yin2024sg} introduces scene graph prompting for hierarchical LLM-based goal reasoning. CoW~\cite{yitzhak2022cows} and OpenFMNav~\cite{kuang2024openfmnav} employ vision-language foundation models for open-vocabulary object search. These methods share an implicit premise: that the VLM possesses visual familiarity with target objects acquired during pre-training. Laboratory instruments such as centrifuges and incubators lie outside this distribution. Whether VLM navigation failure in laboratory settings stems from deficient conceptual knowledge or insufficient visual signal in rendered assets remains an open question.

\section{BioVLN Platform}

\subsection{Overview}

BioVLN operates through a dual-entry pipeline (Fig.~\ref{fig:pipeline}). Scenes can be constructed through two complementary paths. P1 imports designer-authored GLB scenes annotated via the LSAT Blender addon, which extracts goal positions from artist-placed markers and maps them to instrument categories. P2 procedurally generates scenes from an instrument catalog and layout templates, driving a headless Blender builder followed by semantic annotation, navigation mesh generation, and episode construction. Both paths produce identical outputs: a scene GLB with navigation mesh and episodes. Six navigation methods are evaluated in a shared action space, with safety metrics (MCR, VRT) computed from trajectories against the three-zone model of every instrument in the scene. The platform also provides Gym-compatible and trajectory-recording interfaces for policy learning.

\subsection{Problem Formulation}

We describe the platform's core components: the operational-face goal formulation, the laboratory asset catalog, and the three-zone operational envelope model.

\paragraph{Operational-Face Goal Navigation.}
Let $\mathcal{I} = \{I_1, \ldots, I_K\}$ be the set of instruments in a scene. Each instrument $I_i$ is characterized by a 3D centroid $\mathbf{c}_i$, an operational face direction $\mathbf{f}_i \in \mathbb{R}^2$ (unit vector indicating the interactive surface normal), and a half-depth $d_i$ (distance from centroid to face). The goal position is:

\begin{equation}
\mathbf{g}_i = \mathbf{c}_i + \mathbf{f}_i \cdot (d_i + \delta_i)
\label{eq:goal}
\end{equation}
where $\delta_i$ is an instrument-specific approach distance (0.5--0.9\,m). The goal is placed at floor height ($z{=}0$ in Blender Z-up coordinates).

The agent initializes at a random navigable start position $\mathbf{s}$ on the navigation mesh. At each step $t$, the agent receives an egocentric RGB-D observation and selects $a_t \in \mathcal{A} = \{\text{forward } 0.25\text{m}, \text{turn\_left } 30^\circ, \text{turn\_right } 30^\circ, \text{stop}\}$. An episode succeeds if the agent calls \texttt{stop} within a distance $r_i$ of $\mathbf{g}_i$, where $r_i$ is an instrument-specific success radius (0.8--1.3\,m). The task is zero-shot: the agent receives no training episodes from the target scene.

\paragraph{Laboratory Asset Catalog.}
Table~\ref{tab:instruments} lists the instrument and infrastructure assets used in BioVLN. Each asset is stored as a GLB file with geometry centered at the model-space origin, allowing uniform world-space placement and rotation without per-asset coordinate adjustments. Dimensions and support surface heights were verified through Blender vertex analysis. Operational face directions were determined by manual inspection of each instrument's 3D model in Blender.

\begin{table}[t]
\centering
\caption{Laboratory Asset Catalog}
\label{tab:instruments}
{\small
\setlength{\tabcolsep}{2.5pt}
\begin{tabular*}{\columnwidth}{@{\extracolsep{\fill}}c c c c c@{}}
\toprule
Asset & Dimensions (m) & Zone & Op. Face & $d_{\text{half}}$ (m) \\
\midrule
Refrigerator    & $0.89{\times}0.91{\times}1.78$ & wall      & $-Y$ & 0.455 \\
Incubator       & $1.35{\times}0.85{\times}1.90$ & wall      & $-X$ & 0.673 \\
Centrifuge      & $0.49{\times}0.42{\times}0.63$ & table\_top & $-X$ & 0.245 \\
Mini-Centrifuge & $0.66{\times}0.50{\times}0.43$ & table\_top & $-X$ & 0.332 \\
Pipette         & $0.04{\times}0.32{\times}0.28$ & table\_top & $-Y$ & 0.159 \\
Cabinet         & $1.38{\times}0.57{\times}1.77$ & wall      & $+X$ & 0.692 \\
Waste Bin       & $0.38{\times}0.37{\times}0.51$ & floor     & $+X$ & 0.183 \\
\cmidrule{1-5}
\multicolumn{5}{@{}l}{Room infrastructure} \\
\addlinespace
Lab Bench       & $1.55{\times}5.94{\times}0.76$  & --- & --- & --- \\
Lab Table       & $2.40{\times}0.90{\times}1.10$  & --- & --- & --- \\
Room (inner)    & $8.00{\times}6.00{\times}3.24$  & --- & --- & --- \\
\bottomrule
\end{tabular*}}
\end{table}

\begin{figure}[t]
\centering
\includegraphics[width=\columnwidth]{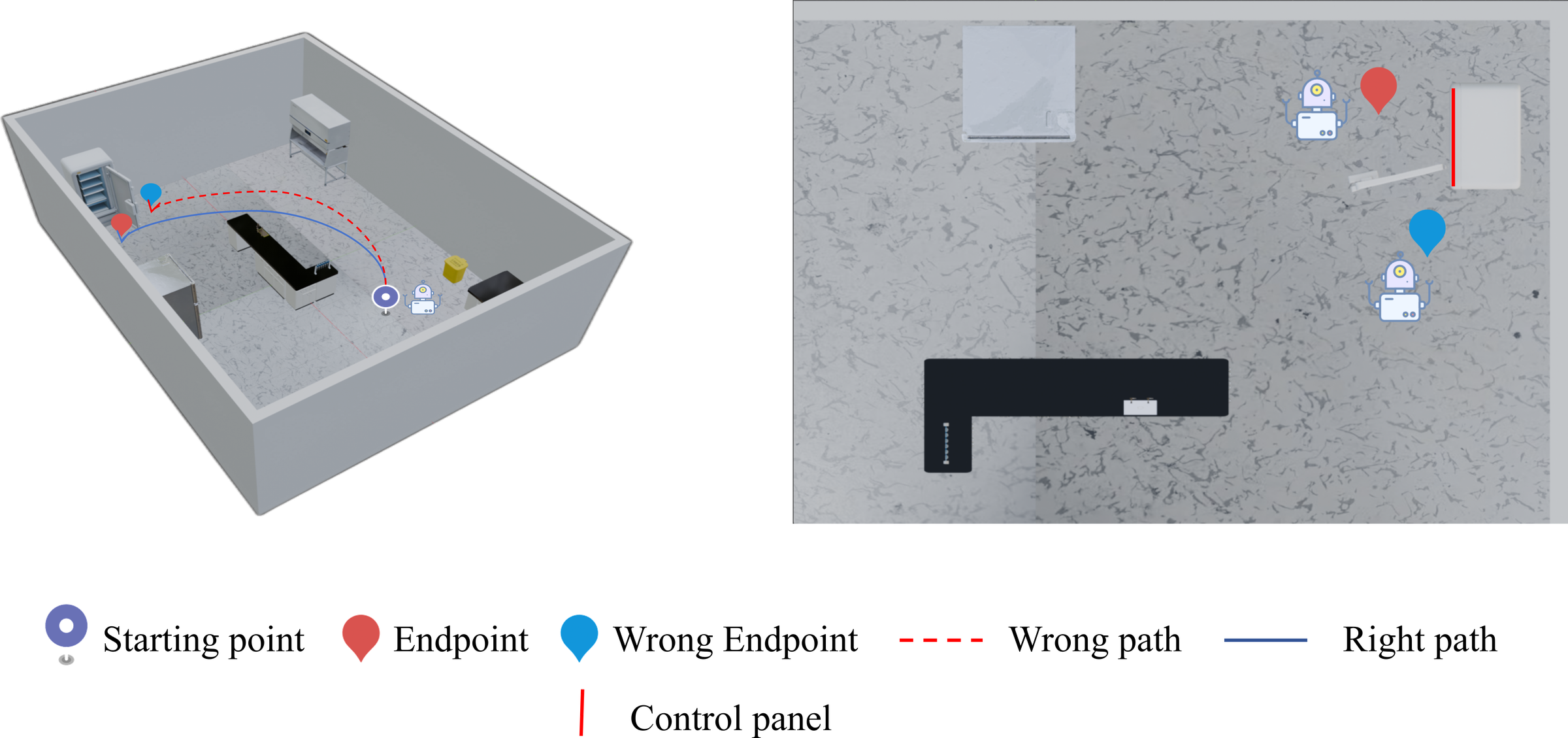}
\caption{Operational-face goal navigation. Left: blue solid = correct navigation to the operational face (red point); red dashed = incorrect (reaches incubator but not its operational surface). Right: top-down view of the operational face.}
\label{fig:three_zone}
\end{figure}

\paragraph{Three-Zone Operational Envelope.}
BioVLN's defining spatial abstraction is a three-zone model around each instrument. The operational-face navigation concept underlying this model is illustrated in Fig.~\ref{fig:three_zone}, and the three zones jointly govern scene generation, collision avoidance, and navigation evaluation.

(1) \textbf{Zone 1: Physical body.} The instrument's axis-aligned bounding box, derived from its GLB geometry (Table~\ref{tab:instruments}). This zone represents the space that the instrument physically occupies. Wall instruments contact the wall on one face; table-top instruments rest on support surfaces at precisely measured heights ($Z_{\text{bench}}{=}0.76$\,m, $Z_{\text{table}}{=}0.78$\,m).

(2) \textbf{Zone 2: Safety clearance.} A buffer of $\varepsilon_c{=}0.25$\,m extends beyond Zone~1 on all four horizontal sides, representing the minimum clearance required for safe navigation. During scene generation, the Zone~2 regions of any two instruments cannot overlap:
\begin{equation}
|x_i - x_j| \geq \tfrac{s_{x,i} + s_{x,j}}{2} + 2\varepsilon_c \;\; \text{or} \;\; |y_i - y_j| \geq \tfrac{s_{y,i} + s_{y,j}}{2} + 2\varepsilon_c
\label{eq:clearance}
\end{equation}
This guarantees navigable gaps of at least $0.5$\,m between instruments. During evaluation, Zone~2 violations are recorded and reported as safety metrics (Fig.~\ref{fig:safety}).

(3) \textbf{Zone 3: Operational envelope.} A rectangular region of depth $\delta_i$ (0.5--0.9\,m, instrument-dependent) extending outward from the operational face along $\mathbf{f}_i$. The goal $\mathbf{g}_i$ sits at its outer boundary (Eq.~\ref{eq:goal}). Zone~3 defines the task-relevant navigation target: the standing position from which a human or robot can operate the instrument.

The three-zone model departs deliberately from standard ObjectNav. Habitat ObjectNav~\cite{savva2019habitat} projects bounding-box centers to the floor, producing positions that may lie inside furniture or against walls. Matterport3D~\cite{chang2017matterport3d} uses human-annotated viewpoints without explicit spatial constraints. BioVLN makes spatial semantics explicit and uses the three-zone model for safety evaluation.

\subsection{Scene Generation Pipeline}

\subsubsection{Zero-Origin Architecture}

Our pipeline uses origin-centered GLB assets: each model's geometry is centered at $(0,0,0)$ in its local coordinate frame, and world-space placement and rotation are applied uniformly at import time. This design contrasts with conventional baked-GLB workflows, where vertex coordinates are pre-transformed to world space, preventing repositioning or layout variation.

P1 (designer-authored import) begins with a manually built GLB scene containing artist-placed \texttt{target\_*} markers. LSAT extracts these markers and maps them to instrument categories. A hand-authored configuration file supplies semantic mesh annotations, after which the shared pipeline stages (semantic annotation, navigation mesh, episode construction) produce the final benchmark scene.

P2 (procedural generation) uses a specification generator that produces a JSON description encoding each instrument's world position, rotation, and operational goal. A headless Blender builder imports room and instrument GLBs at the specified positions, then the shared pipeline stages process the assembled scene.

Both paths converge on identical outputs: a scene GLB file with vertex-level semantic annotations (COLOR\_0 encoding), a Recast navigation mesh, and episode definitions with sampled starts and operational-face goals.

\subsubsection{Rotation and Placement}

Instruments against walls must rotate so their operational face points toward the room interior. Given a model-space face vector $(m_x, m_y)$ and wall-determined interior direction $(w_x, w_y)$, the Z-axis rotation $\theta$ is:

\begin{equation}
\theta = \text{atan2}(m_x w_y - m_y w_x,\; m_x w_x + m_y w_y)
\label{eq:rotation}
\end{equation}

This formulation handles all four wall orientations and arbitrary model-space face directions. Table-top instruments align perpendicular to the nearest table edge; floor instruments (waste bin) are unconstrained.

Vertical placement uses surface height and model offsets:
\begin{equation}
Z_{\text{place}} = Z_{\text{table\_top}} + Z_{\text{bb\_center}} - Z_{\text{bottom}}
\label{eq:placement}
\end{equation}
where $Z_{\text{table\_top}}$ is the measured support surface height (0.76\,m bench, 0.78\,m table), $Z_{\text{bb\_center}}$ is the instrument's bounding-box center Z, and $Z_{\text{bottom}}$ is its lowest vertex Z.

\subsubsection{Layout Templates}

\begin{figure}[t]
\centering
\includegraphics[width=\columnwidth]{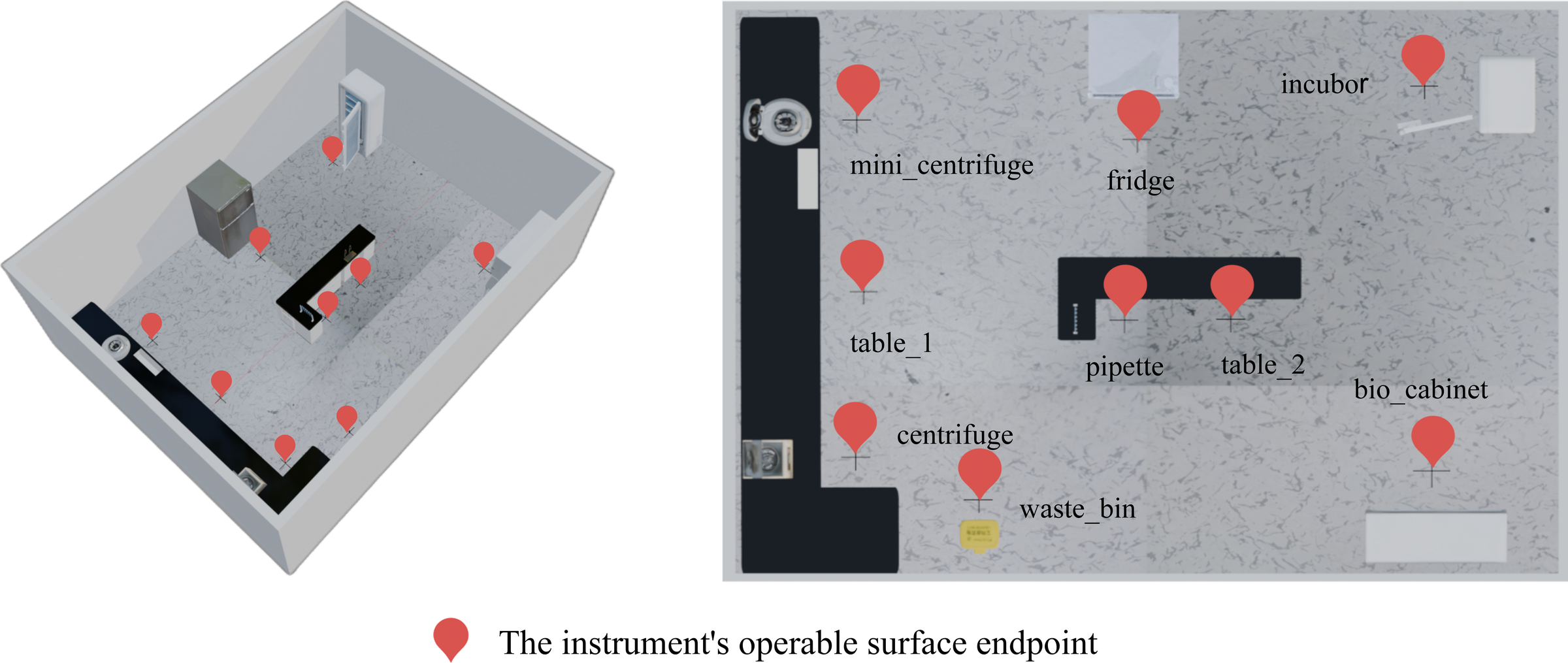}
\caption{Scene-level operational-face goal positions. Left: oblique overview. Right: top-down visualization per instrument.}
\label{fig:scene_layout}
\end{figure}

We define four layout templates; two (single-room, two-room) are implemented and used in benchmarks, while suite and open-plan are defined for future expansion. Fig.~\ref{fig:scene_layout} shows the resulting goal positions across all instruments in a single-room scene. Table~\ref{tab:templates} summarizes generation statistics.

\begin{table}[t]
\centering
\caption{Layout Templates and Generation Statistics}
\label{tab:templates}
\begin{tabular*}{\columnwidth}{@{\extracolsep{\fill}}c c c c@{}}
\toprule
Template & Room Configuration & Scenes & Episodes \\
\midrule
single-room & $8{\times}6$\,m                      & 39 & 1{,}363 \\
two-room    & $6{\times}5$\,m + $3.5{\times}3.5$\,m  & 8  & 304 \\
\midrule
Total & & 47 & 1{,}667 \\
\bottomrule
\end{tabular*}
\end{table}

The generator randomizes which instrument slots are active (2--5 out of 6 for single-room) and which instruments fill those slots, producing layout diversity. Each scene is identified by a seed value that deterministically controls random choices.

\subsubsection{Episode Construction}

Start positions are randomly sampled from the navigation mesh at least 1.0\,m from obstacles. Goal positions are snapped to the nearest navigable navmesh point. The geodesic shortest path from start to goal is computed via the navmesh graph. View-points are generated at the goal, oriented toward the target instrument. Episodes are classified by geodesic distance: easy ($d_{\text{geo}} < 3.0$\,m), medium ($3.0 \le d_{\text{geo}} \le 6.0$\,m), hard ($d_{\text{geo}} > 6.0$\,m). On Multi-Scene DEV (160 episodes), the distribution is 11.9\% easy, 28.1\% medium, 60.0\% hard.

\subsection{LSAT Annotation Toolkit}

We developed the LabScene Annotation Toolkit (LSAT), a Blender 5.1.2 addon (13 modules, $\sim$2{,}500 lines of Python) that bridges designer-authored 3D scenes and the BioVLN benchmark format. LSAT implements two complementary goal-extraction strategies. The first extracts world-space coordinates from artist-placed \texttt{target\_*} empty nodes in the GLTF file, mapping node names to instrument categories through fuzzy string matching (Levenshtein distance threshold of 2). The second computes goal positions from detected instrument meshes using their centroids and operational-face geometry (Eq.~\ref{eq:goal}), serving as a fallback when target markers are absent. LSAT achieved 100\% coverage on the human-annotated test scene and exports to BioVLN goal format, LSAT internal format, and Isaac Sim USD.

\section{Experimental Setup}

\subsection{Benchmark Datasets}

We define three benchmark datasets (Table~\ref{tab:datasets}):

\begin{table}[t]
\centering
\caption{Benchmark Dataset Statistics}
\label{tab:datasets}
{\small
\setlength{\tabcolsep}{1.5pt}
\begin{tabular*}{\columnwidth}{@{\extracolsep{\fill}}c c c c c@{}}
\toprule
Dataset & Scenes & Eps. & Cat. & Goal Source \\
\midrule
MS DEV  & 4 & 160 & 4  & Computed \\
MS VAL  & 1 & 50  & 4  & Computed \\
MS TEST & 2 & 120 & 4  & Computed \\
Two-Room         & 1  & 50  & 10 & Computed \\
LSAT Target      & 1  & 40  & 8  & Human-annotated \\
\midrule
Total   & --- & 420 & --- & \\
\bottomrule
\end{tabular*}}
\end{table}

Multi-Scene spans 7 held-out scenes (4 DEV, 1 VAL, 2 TEST) drawn from 47 total built scenes across 2 layout templates, with 2--4 instrument categories per scene. The three-way split ensures no scene appears in more than one split, testing generalization. Two-Room contains two connected rooms joined by a 1.5\,m-wide central corridor, with 10 instrument categories and 50 episodes specifically testing cross-room navigation. LSAT Target is a human-annotated scene with 8 instrument categories and LSAT Strategy~1 goal positions, isolating the effect of goal placement precision.

\subsection{Evaluation Metrics}

We report six metrics spanning navigation effectiveness and operational safety.

Navigation Metrics. (1) Success Rate (SR): fraction of episodes ending within the instrument-specific success radius. (2) SPL: success weighted by $\ell_i^{\text{opt}} / \max(\ell_i, \ell_i^{\text{opt}})$, where $\ell_i^{\text{opt}}$ is the optimal geodesic path length. (3) Distance to Goal (DTG): final Euclidean distance to the goal, averaged over all episodes. (4) Average Steps: mean actions per successful episode.

Safety Metrics. Two metrics derived from the three-zone model, computed from the agent's full trajectory against all instruments in the scene: (1) Minimum Clearance Radius (MCR): the closest distance (m) between the agent and any instrument surface over the entire trajectory. Values below the Zone~2 hazard threshold ($0.5$\,m) indicate a safety violation. (2) Violation Rate at Threshold (VRT): the fraction of trajectory steps during which the agent is within $0.5$\,m of any instrument surface. VRT captures persistent unsafe proximity, not single-point violations.

\subsection{Baseline Methods}

Six methods spanning random actions to VLM-guided semantic exploration, all operating in the same action space $\mathcal{A}$ with a 500-step maximum.

Random. Uniformly samples from the three movement actions (forward, turn left, turn right) with a 5\% per-step stop probability. Serves as a pure-chance lower bound.

Oracle. Follows the pre-computed geodesic shortest path on the navigation mesh. Because navmesh-snapped goals may be unreachable under discrete actions (0.25\,m forward, $30^\circ$ turns), the Oracle does not achieve 100\% success; it provides an empirical upper bound.

Frontier Exploration. A purely geometric method using a depth-based 2D occupancy grid ($0.05$\,m resolution, $100{\times}100$\,m extent), frontier clustering, and nearest-frontier navigation. This baseline tests whether systematic geometric coverage alone suffices for laboratory navigation, without any semantic understanding of instrument appearance or location.

LLM-Frontier. Extends Frontier Exploration by querying DeepSeek-chat~\cite{xu2026deepseek} when multiple frontiers are approximately equidistant. The LLM receives a structured prompt describing the target instrument, agent position, and top-3 frontier candidates, then selects which to pursue. This baseline tests whether an LLM's semantic priors about instrument locations improve frontier selection in laboratory layouts.

3-Zone Oracle. An oracle agent that samples multiple candidate approach points within the target instrument's Zone~3, navigating to each until one succeeds. This baseline isolates the benefit of zone-aware multi-point goal sampling over single-point navmesh snapping.

VLFM. Our reimplementation of VLFM~\cite{yokoyama2024vlfm} with \texttt{vlm\_vmap} mode: Grounding DINO~\cite{liu2024grounding} for open-vocabulary detection, GPT-4o~\cite{hurst2024gpt} for detection verification and direction scoring, and SAM~\cite{kirillov2023segment} for segmentation. The agent early-stops after 100 consecutive steps without progress. VLFM tests whether pretrained visual representations from VLM-guided zero-shot navigation transfer to rendered laboratory assets.

\subsection{Implementation Details}

All experiments run on a single NVIDIA RTX 4090 (24\,GB) with habitat-sim 0.2.5. Frontier Exploration and LLM-Frontier operate at $\sim$5 steps/s; VLFM at $\sim$0.25 steps/s due to VLM API latency. GPT-4o and DeepSeek are accessed via API endpoints. LSAT is developed for Blender 5.1.2~\cite{community2018blender}.

Beyond zero-shot evaluation, BioVLN supports trajectory collection and policy learning through a trajectory recorder that captures per-step observations during evaluation, a PyTorch dataset class that computes agent-centric goal vectors, and a Gymnasium environment with step-wise reward:
\begin{equation}
\begin{aligned}
r_t =\;& (d_{t-1} - d_t) \;+\; \mathbb{I}[d_t < r_i] \cdot 10 \\
      &-\; 0.01 \;-\; \mathbb{I}[d_{\text{clearance}} < 0.5] \cdot 0.05
\end{aligned}
\label{eq:reward}
\end{equation}
where $d_t$ is the Euclidean distance to goal at step $t$, $r_i$ is the instrument-specific success radius, and $d_{\text{clearance}}$ is the minimum distance to any instrument surface. The first term provides dense reward for approaching the goal; the second awards a sparse success bonus; the step penalty encourages efficient paths; the safety term penalizes proximity below the Zone~2 threshold.

\section{Experiments}

\begin{table*}[t]
\centering
\caption{Navigation and safety results across all benchmarks.}
\label{tab:results}
{\footnotesize
\setlength{\tabcolsep}{0pt}
\begin{tabular*}{\textwidth}{@{\extracolsep{\fill}}l c c c c c c c c c c c c c c c@{}}
\toprule
& \multicolumn{4}{c}{MS DEV} & \multicolumn{2}{c}{MS VAL} & \multicolumn{2}{c}{MS TEST} & \multicolumn{4}{c}{Two-Room} & \multicolumn{3}{c@{}}{LSAT} \\
\cmidrule(lr){2-5} \cmidrule(lr){6-7} \cmidrule(lr){8-9} \cmidrule(lr){10-13} \cmidrule(lr){14-16}
Method & \ww{SR} & \ww{SPL} & \ww{MCR} & \ww{VRT} & \ww{SR} & \ww{SPL} & \ww{SR} & \ww{SPL} & \ww{SR} & \ww{SPL} & \ww{MCR} & \ww{VRT} & \ww{SR} & \ww{MCR} & \ww{VRT} \\
\midrule
Random    & \ww{26.9} & \ww{0.212} & \ww{0.976} & \ww{11.9} & \ww{22.0} & \ww{0.149} & \ww{31.7} & \ww{0.270} & \ww{10.0} & \ww{0.066} & \ww{0.637} & \ww{12.7} & \ww{32.5} & \ww{0.135} & \ww{55.3} \\
Oracle    & \ww{71.9} & \ww{0.719} & \ww{0.944} & \ww{9.7}  & \ww{76.0} & \ww{0.760} & \ww{79.2} & \ww{0.792} & \ww{80.0} & \ww{0.800} & \ww{0.563} & \ww{10.7} & \ww{77.5} & \ww{0.124} & \ww{68.2} \\
Frontier  & \ww{74.4} & \ww{0.744} & \ww{0.918} & \ww{9.8}  & \ww{80.0} & \ww{0.800} & \ww{82.5} & \ww{0.825} & \ww{84.0} & \ww{0.840} & \ww{0.504} & \ww{11.3} & \ww{87.5} & \ww{0.104} & \ww{70.3} \\
LLM-Frn. & \ww{74.4} & \ww{0.744} & \ww{---}  & \ww{---}  & \ww{84.0} & \ww{0.839} & \ww{82.5} & \ww{0.825} & \ww{84.0} & \ww{0.840} & \ww{---}  & \ww{---}  & \ww{87.5} & \ww{---}  & \ww{---}  \\
3-Zone Or.& \ww{83.8} & \ww{0.838} & \ww{1.115} & \ww{4.9}  & \ww{84.0} & \ww{0.840} & \ww{83.3} & \ww{0.833} & \ww{86.0} & \ww{0.860} & \ww{0.672} & \ww{6.3}  & \ww{92.5} & \ww{0.110} & \ww{61.3} \\
VLFM      & \ww{19.4} & \ww{---}  & \ww{1.853} & \ww{4.7}  & \ww{38.0} & \ww{---}  & \ww{---}  & \ww{---}  & \ww{28.0} & \ww{---}  & \ww{0.822} & \ww{3.8}  & \ww{62.5} & \ww{2.165} & \ww{5.3}  \\
\bottomrule
\multicolumn{16}{@{}l}{\footnotesize SR: success rate (\%). SPL: success weighted by path length. MCR: min.\ clearance radius (m). VRT: violation rate (\%).} \\
\multicolumn{16}{@{}l}{\footnotesize VLFM SPL N/A (VLM paths not geodesic-comparable). TEST omitted (API limits). MS-VAL safety: MCR=1.07, VRT=18.2.} \\
\multicolumn{16}{@{}l}{\footnotesize LLM-Frontier safety: --- = identical to Frontier (same geometric path).} \\
\end{tabular*}}
\end{table*}

\subsection{Geometric Coverage Is Sufficient for Single-Room Layouts}

Table~\ref{tab:results} reports all metrics. Frontier Exploration achieves 74.4\% SR on MS-DEV, surpassing the geodesic Oracle (71.9\%) by 2.5 pp (McNemar $p{=}0.219$); on Two-Room, Frontier reaches 84.0\% vs.\ Oracle's 80.0\% ($+4.0$\,pp, $p{=}0.500$). Systematic geometric coverage compensates for discretization error in reaching navmesh-snapped goals under 0.25\,m steps and $30^\circ$ turns. 3-Zone Oracle achieves the highest SR overall (83.3--92.5\%), confirming that zone-aware multi-point sampling further compensates for discretization. LLM-Frontier matches Frontier on 4 of 5 splits; the LLM is consulted in only 10--20\% of episodes because single-room layouts lack room-level semantic choices. VLFM achieves 62.5\% SR on LSAT Target but drops to 19.4\% on MS-DEV; on Two-Room, all successes are confined to Room~A instruments.

\subsection{Per-Instrument Difficulty Reflects Spatial Layout}

Table~\ref{tab:per_category} disaggregates Frontier Exploration by instrument category. Wall instruments (refrigerator: 60--65\%; cabinet: 70--100\%) are harder than table-top instruments (centrifuge: 80--100\%), whose approach paths benefit from open space at the room center. LSAT Target's higher SR (87.5\%) confirms that human annotators avoid placing goals near obstacles.

\begin{table}[t]
\centering
\caption{Frontier SR by Instrument Category (\%)}
\label{tab:per_category}
{\small
\begin{tabular*}{\columnwidth}{@{\extracolsep{\fill}}c c c c@{}}
\toprule
Category & MS-DEV & LSAT & Two-Room \\
\midrule
Cabinet         & 70.0  & 80.0  & 100.0 \\
Centrifuge      & 80.0  & ---   & 100.0 \\
Incubator       & ---   & 100.0 & 80.0 \\
Lab Bench       & ---   & 100.0 & 100.0 \\
Lab Table       & ---   & 80.0  & 60.0 \\
Mini-Centrifuge & ---   & 100.0 & 80.0 \\
Pipette         & ---   & 60.0  & 80.0 \\
Refrigerator    & 65.0  & 80.0  & 60.0 \\
Waste Bin       & 82.5  & 100.0 & 80.0 \\
\midrule
Overall & 74.4  & 87.5  & 84.0 \\
\bottomrule
\multicolumn{4}{@{}l}{\footnotesize --- : instrument category not present in that dataset.} \\
\end{tabular*}}
\end{table}

\subsection{Safety Depends on Layout Density}

\begin{figure}[t]
\centering
\includegraphics[width=\columnwidth]{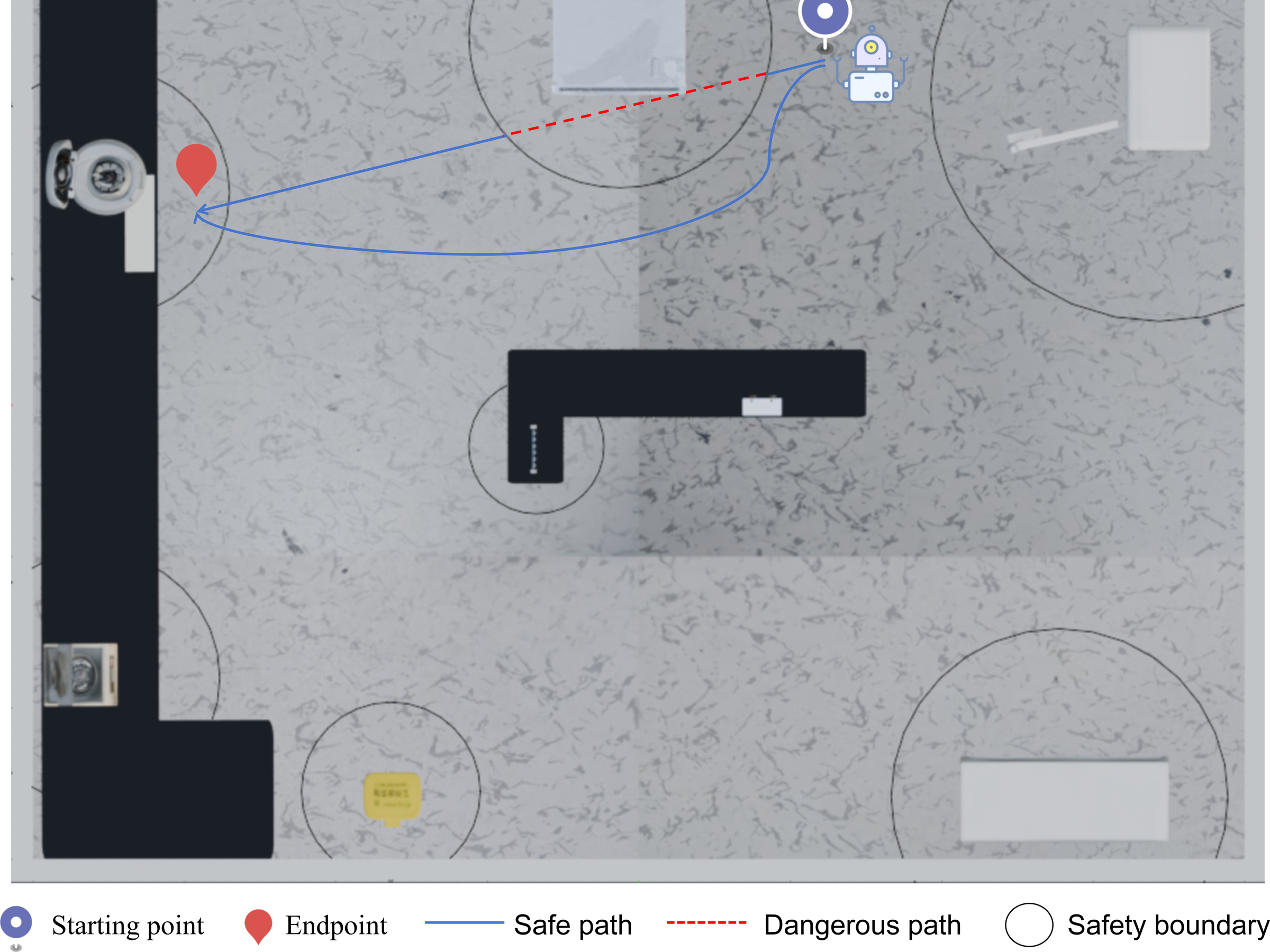}
\caption{Safety boundary: red dashed violates instrument clearance; blue solid maintains safe clearance.}
\label{fig:safety}
\end{figure}

Fig.~\ref{fig:safety} shows an example trajectory with safety violations. Safety is dominated by layout density (Table~\ref{tab:results}). On the compact LSAT layout, VRT reaches 55.3--70.3\% for Frontier and Oracle, spending more than half the episode within 0.5\,m of instruments. On Two-Room, VRT drops to 6.3--11.3\% because the corridor separates navigation paths from instrument zones. VLFM achieves the lowest VRT but also the lowest SR, revealing a success-safety trade-off. 3-Zone Oracle achieves the lowest VRT among high-SR methods by constraining navigation endpoints to Zone~3.

\subsection{VLM Analysis}

GPT-4o identifies all instruments from text yet only 13.3\% from rendered views (Table~\ref{tab:vlm_gap}). Recognition follows surface color diversity: 3/3 for the biosafety cabinet (12.6k--14.9k colors), 0/6 for near-uniform surfaces. VLM failure stems from sparse textures, not deficient knowledge.

\begin{table}[t]
\centering
\caption{VLM Recognition vs.\ Surface Color Diversity}
\label{tab:vlm_gap}
{\small
\setlength{\tabcolsep}{2pt}
\begin{tabular}{@{}l c c c c@{}}
\toprule
Instrument & Colors & GPT-4o Answer & Correct & $n$ \\
\midrule
Biosafety Cab. & 12.6k--14.9k & ``Biosafety cabinet'' & 3/3 & 3 \\
Incubator       & 6.1k--7.6k   & ``Laboratory freezer'' & 0/3 & 3 \\
Centrifuge      & 1.9k--14.0k  & ``Yellow trash bin''   & 0/6 & 6 \\
Micropipette    & 1.8k--19.8k  & ``Lab equipment''       & 1/6 & 6 \\
Refrigerator    & 1--1.8k      & ``Gray rectangle'' & 0/6 & 6 \\
Waste Bin       & 1--2         & ``Gray square'' & 0/6 & 6 \\
\bottomrule
\end{tabular}}
\end{table}

\subsection{Operational-Face Goals Are Layout-Invariant}

Frontier Exploration with centroid vs.\ operational-face goals on the combined 330-episode Multi-Scene set yields identical aggregate SR (78.2\%, Table~\ref{tab:ablation}). Aggregate parity masks opposing per-split effects: centroid improves DEV (+8.1 pp, $p{=}0.035$) yet degrades VAL ($-$18.0 pp, $p{=}0.035$). Each formulation wins 38 episodes the other loses. Operational-face goals avoid this layout-dependent fluctuation because the goal is defined relative to the oriented face, not the centroid projection.

\begin{table}[t]
\centering
\caption{Centroid vs.\ Operational-Face Goals}
\label{tab:ablation}
{\small
\begin{tabular}{@{}l c c c c@{}}
\toprule
Split ($n$) & Centroid & Op-Face & $\Delta$ & $p$ \\
\midrule
DEV (160)    & 82.5 & 74.4 & +8.1  & 0.035 \\
VAL (50)     & 62.0 & 80.0 & $-$18.0 & 0.035 \\
TEST (120)   & 79.2 & 82.5 & $-$3.3  & 0.557 \\
\midrule
Overall & 78.2 & 78.2 & 0.0 & --- \\
\bottomrule
\end{tabular}}
\end{table}

\subsection{Behavioral Cloning Baseline}

A behavioral cloning policy trained on Frontier trajectories achieves 41.9\% SR on held-out VAL (vs.\ Frontier's 80.0\%), establishing a reference for future learned methods.

\section{Conclusion}

We presented BioVLN, a simulation platform for biomedical laboratory navigation built on operational-face goals and a three-zone operational envelope. Frontier Exploration (74.4--87.5\% SR) approaches the oracle bound through geometric coverage. 3-Zone Oracle (83.3--92.5\%) achieves the highest success by sampling within the operational envelope. VLFM (19.4--62.5\%) reveals a surface-texture bottleneck: GPT-4o identifies all instruments from text but recognizes only 13.3\% from rendered views. A behavioral cloning baseline (41.9\% SR) confirms the training pipeline.

Key limitations and future directions:

\textbf{Asset visual quality.} Flat-shaded materials limit VLM recognition; richer textures would bridge this domain gap.

\textbf{Scene diversity.} Texture randomization and clutter would produce more realistic laboratory scenes.

\textbf{Domain generalization.} The operational-face model extends to any domain with affordance-directed surfaces.

\textbf{Sim-to-real transfer.} Deploying trained policies to physical robots remains an open direction.

\bibliographystyle{references}
\bibliography{references}

\appendix
\section{Per-Method Detailed Metrics}

Table~\ref{tab:appendix_detailed_a} reports Success Rate (SR, \%), Distance to Goal (DTG, m), and average steps per successful episode for all methods evaluated on the Multi-Scene dataset. DTG measures the Euclidean distance between the agent's final position and the operational-face goal. Average steps are computed over successful episodes only.

\begin{table*}[t]
\centering
\caption{Detailed Navigation Metrics on Multi-Scene}
\label{tab:appendix_detailed_a}
{\small
\begin{tabular*}{\textwidth}{@{\extracolsep{\fill}}l c c c c c c c c c@{}}
\toprule
& \multicolumn{3}{c}{DEV} & \multicolumn{3}{c}{VAL} & \multicolumn{3}{c@{}}{TEST} \\
\cmidrule(lr){2-4} \cmidrule(lr){5-7} \cmidrule(l){8-10}
Method & SR & DTG & Steps & SR & DTG & Steps & SR & DTG & Steps \\
\midrule
Random    & 26.9 & 5.20 & 410.1 & 22.0 & 4.03 & 426.9 & 31.7 & 3.99 & 382.3 \\
Oracle    & 71.9 & 2.96 & 157.1 & 76.0 & 2.31 & 133.1 & 79.2 & 2.75 & 117.3 \\
Frontier  & 74.4 & 2.98 & 145.5 & 80.0 & 2.15 & 114.5 & 82.5 & 2.82 & 101.5 \\
LLM-Frontier & 74.4 & 2.97 & 145.5 & 84.0 & 2.43 & 105.5 & 82.5 & 2.82 & 101.5 \\
3-Zone Oracle & 83.8 & 3.87 & 97.0  & 84.0 & 2.85 & 94.7  & 83.3 & 3.15 & 95.6  \\
\bottomrule
\end{tabular*}}
\end{table*}

Table~\ref{tab:appendix_detailed_b} reports metrics for LSAT Target and Two-Room datasets. Due to differences in evaluation infrastructure, per-episode DTG and step counts are not available for the Random and Oracle baselines on these datasets; wall-clock elapsed time is reported instead.

\begin{table*}[t]
\centering
\caption{Detailed Metrics on LSAT Target and Two-Room}
\label{tab:appendix_detailed_b}
{\small
\begin{tabular*}{\textwidth}{@{\extracolsep{\fill}}l c c c c c c@{}}
\toprule
& \multicolumn{3}{c}{LSAT Target (40 eps)} & \multicolumn{3}{c@{}}{Two-Room (50 eps)} \\
\cmidrule(lr){2-4} \cmidrule(l){5-7}
Method & SR & Elapsed (s) & SPL & SR & Elapsed (s) & SPL \\
\midrule
Random    & 32.5 & 1.6  & ---  & 10.0 & 2.1  & ---  \\
Oracle    & 77.5 & 1.5  & 0.775 & 80.0 & 1.7  & 0.800 \\
Frontier  & 87.5 & 7.4  & 0.875 & 84.0 & 14.1 & 0.840 \\
3-Zone Oracle & 92.5 & 1.5  & 0.925 & 86.0 & 1.6  & 0.860 \\
\bottomrule
\end{tabular*}}
\end{table*}

\section{Per-Difficulty Performance}

Episodes are stratified by geodesic distance $d_{\text{geo}}$: easy ($d_{\text{geo}} < 3.0$\,m), medium ($3.0 \le d_{\text{geo}} \le 6.0$\,m), and hard ($d_{\text{geo}} > 6.0$\,m). On Multi-Scene DEV (160 episodes), the class distribution is 11.9\% easy, 28.1\% medium, and 60.0\% hard. Table~\ref{tab:appendix_difficulty} reports Frontier and 3-Zone Oracle SR disaggregated by difficulty across Multi-Scene splits.

\begin{table}[t]
\centering
\caption{SR (\%) by Difficulty Tier}
\label{tab:appendix_difficulty}
{\small
\begin{tabular*}{\columnwidth}{@{\extracolsep{\fill}}l c c c c c c@{}}
\toprule
& \multicolumn{3}{c}{Frontier} & \multicolumn{3}{c@{}}{3-Zone Oracle} \\
\cmidrule(lr){2-4} \cmidrule(l){5-7}
Difficulty & DEV & VAL & TEST & DEV & VAL & TEST \\
\midrule
Easy   & 89.5 & 100.0 & 100.0 & 100.0 & 100.0 & 100.0 \\
Medium & 84.4 & 73.7  & 83.3  & 97.8  & 73.7  & 83.3  \\
Hard   & 66.7 & 70.6  & 68.9  & 74.0  & 82.3  & 71.1  \\
\bottomrule
\end{tabular*}}
\end{table}

Both methods achieve uniformly high SR for easy episodes. For hard episodes, performance drops to 66.7--82.3\%, consistent with the expectation that geodesic distance is the primary source of navigation difficulty in single-room laboratory layouts.

\section{Centroid vs.\ Operational-Face Ablation}

Frontier Exploration was evaluated under two goal formulations on Multi-Scene (330 episodes). In the centroid condition, goals are defined as the floor projection of the instrument bounding-box center, consistent with standard ObjectNav. In the operational-face condition, goals lie at the outer boundary of Zone~3 (Eq.~1, main text). McNemar's test for paired binary outcomes was applied per split.

\begin{table*}[t]
\centering
\caption{Centroid vs.\ Operational-Face Goals --- Per-Split Comparison}
\label{tab:appendix_ablation}
{\small
\begin{tabular*}{\textwidth}{@{\extracolsep{\fill}}l c c c c c c@{}}
\toprule
Split ($n$) & Centroid SR & Op-Face SR & $\Delta$ (pp) & $p$ (McNemar) & Cent.\ Wins & OpF.\ Wins \\
\midrule
DEV (160)   & 82.5 & 74.4 & +8.1  & 0.035 & 28 & 15 \\
VAL (50)    & 62.0 & 80.0 & $-$18.0 & 0.035 & 3  & 12 \\
TEST (120)  & 79.2 & 82.5 & $-$3.3  & 0.557 & 7  & 11 \\
\midrule
Overall (330) & 78.2 & 78.2 & 0.0 & ---  & 38 & 38 \\
\bottomrule
\multicolumn{7}{@{}l}{\footnotesize Cent.\ Wins (OpF.\ Wins): episodes where centroid (operational-face) formulation succeeds and the other fails.} \\
\end{tabular*}}
\end{table*}

The opposing per-split effects correspond to instrument placement. On DEV, centroid goals raise refrigerator SR (85.0\% vs.\ 65.0\%) by removing the approach-direction constraint; the agent can reach the refrigerator from any side. On VAL, centroid goals degrade refrigerator SR (40.0\% vs.\ 100.0\%) because the projected centroid lies behind a wall and is unreachable on the navigation mesh. Across the two splits, the two effects cancel at the aggregate level (78.2\% each). Cabinet and centrifuge exhibit no material difference between conditions, as their placement positions project to navigable floor locations in both formulations.

\section{VLM Recognition Experiment Details}

A recognition experiment was conducted to determine whether VLFM navigation failures originate from deficient domain knowledge or insufficient visual texture in rendered simulation frames. Six laboratory instruments were rendered at 1024$\times$768 pixels under the default Habitat-sim \texttt{no\_lights} rendering mode, with no post-processing. For each instrument, 3--6 distinct viewpoints were captured from the operational-face goal position, oriented toward the instrument.

GPT-4o (gpt-4o-2024-08-06) was queried per viewpoint: \textit{``What laboratory instrument is shown in this image? Answer with the instrument name only.''} Unique surface colors were computed by counting distinct RGB values within the instrument's segmented region (background removed via SAM mask).

A text-only control confirmed that GPT-4o identifies all six instruments by name, demonstrating complete conceptual knowledge. Recognition from rendered images follows surface color diversity (Table~8, main text): the biosafety cabinet (12.6k--14.9k unique colors; visible vents, seams, and labels) is correctly identified in 3/3 viewpoints; instruments with near-uniform surfaces (refrigerator: 1--1.8k colors, waste bin: 1--2 colors) are classified as generic geometric shapes in 0/6 viewpoints.

Two rendering factors influence the degree of texture sparsity. First, the \texttt{no\_lights} rendering mode suppresses dynamic lighting, reducing per-instrument color diversity by a factor of 3--5$\times$ compared to lit rendering. Second, gamma correction applied during frame extraction further homogenizes pixel intensity. After correcting the lighting pipeline, recognition improved for instruments with moderate texture (incubator, centrifuge) but not for near-uniform surfaces (refrigerator, waste bin).

\section{McNemar Test Formulation}

McNemar's test for paired nominal data evaluates whether the per-episode success/failure patterns of two methods $(A,B)$ differ in distribution. For a given data split, let $n_{00}$ denote episodes where both methods fail, $n_{01}$ where $A$ fails and $B$ succeeds, $n_{10}$ where $A$ succeeds and $B$ fails, and $n_{11}$ where both succeed. The test statistic is:
\begin{equation}
\chi^2 = \frac{(|n_{01} - n_{10}| - 1)^2}{n_{01} + n_{10}}
\end{equation}
with one degree of freedom. When $n_{01} + n_{10} < 10$, the exact binomial test is used. All reported $p$-values are two-sided.

\begin{table*}[t]
\centering
\caption{Pairwise McNemar Test Results}
\label{tab:appendix_mcnemar}
{\small
\begin{tabular*}{\textwidth}{@{\extracolsep{\fill}}l l c c c c@{}}
\toprule
Method A & Method B & Split & $n_{01}$ & $n_{10}$ & $p$ \\
\midrule
Frontier & Oracle & MS DEV   & 20 & 16 & 0.619 \\
Frontier & Oracle & MS VAL   & 8  & 6  & 0.791 \\
Frontier & Oracle & MS TEST  & 16 & 12 & 0.572 \\
Frontier & Oracle & Two-Room & 5  & 3  & 0.727 \\
LLM-Frontier & Frontier & MS VAL  & 2  & 0  & --- \\
\bottomrule
\multicolumn{6}{@{}l}{\footnotesize $n_{01}$: $A$ fails \& $B$ succeeds. $n_{10}$: $A$ succeeds \& $B$ fails.} \\
\multicolumn{6}{@{}l}{\footnotesize LLM-Frontier vs.\ Frontier on MS VAL: $n_{01}{+}n_{10}{<}10$, test not applicable.} \\
\end{tabular*}}
\end{table*}

\section{Scene Generation and LSAT Architecture}

Table~\ref{tab:appendix_scenes} summarizes the per-template scene generation statistics for the BioVLN platform. The single-room template randomizes 2--5 of 6 instrument slots per scene; the two-room template uses 10 fixed instrument slots across two connected rooms.

\begin{table}[t]
\centering
\caption{Scene Generation by Layout Template}
\label{tab:appendix_scenes}
{\small
\setlength{\tabcolsep}{8pt}
\begin{tabular}{@{}l c c c c@{}}
\toprule
Template & Instr. & Furn. & Active Slots & Episodes \\
\midrule
single\_room & 4  & 3 & 2--5  & 1{,}363 \\
two\_room    & 10 & 3 & 10 & 304 \\
\midrule
Total & --- & --- & --- & 1{,}667 \\
\bottomrule
\multicolumn{5}{@{}l}{\footnotesize single\_room: 35 (2 active), 30 (3), 40 (4--5) eps/scene.} \\
\multicolumn{5}{@{}l}{\footnotesize two\_room: 38 eps/scene, all 10 slots active in every scene.} \\
\end{tabular}}
\end{table}

The LSAT (LabScene Annotation Toolkit) addon comprises 12 Python modules in three layers. The parsing layer handles GLB deserialization, GLTF node traversal, and Blender-to-Habitat coordinate transforms. The extraction layer implements two strategies: Strategy~1 extracts world-space coordinates from artist-placed \texttt{target\_*} empty nodes via fuzzy string matching (Levenshtein distance threshold of 2) and maps matched names to instrument categories; Strategy~2 detects instrument meshes by name matching, computes centroids and bounding-box dimensions, and applies Eq.~1 for goal computation. The export layer generates BioVLN goal JSON, an internal intermediate format, and USD files for Isaac Sim.

When Strategy~2 cannot determine an operational face direction from geometry alone, the most frequent configuration for that instrument category is used as a default (derived from the instrument catalog, Table~2 of the main text).

\section{Per-Method Experiment Configuration}

Table~\ref{tab:appendix_method_config} specifies the complete configuration for each baseline method evaluated in the main text. All methods share the action space $\mathcal{A} = \{\text{forward } 0.25\text{m}, \text{turn\_left } 30^\circ, \text{turn\_right } 30^\circ, \text{stop}\}$ (except BioVLNGym training, which uses $10^\circ$ turns as noted below) and a 500-step episode limit. Habitat-sim rendering is configured at $640\times480$ RGB-D with \texttt{no\_lights} mode; the \texttt{no\_lights} mode flattens lighting but provides consistent color across viewpoints, which is relevant when interpreting VLM recognition results.

\begin{table*}[t]
\centering
\caption{Per-Method Experiment Configuration}
\label{tab:appendix_method_config}
{\small
\begin{tabular*}{\textwidth}{@{\extracolsep{\fill}}l p{3.8cm} p{3.8cm} p{3.8cm}@{}}
\toprule
Method & Core Parameters & External Dependencies & Runtime \\
\midrule
Random      & Stop probability: 5\% per step & None & $<0.01$\,s/step \\
\addlinespace
Oracle      & Navmesh shortest-path follower, goal radius: 0.8--1.3\,m (instru.-specific) & Recast navmesh & $<0.01$\,s/step \\
\addlinespace
Frontier    & Occupancy grid: $0.05$\,m res., $100{\times}100$\,m extent; cluster radius: $1.0$\,m; min frontier size: 10 cells; depth clip: $10$\,m & None (geometry-only) & $\sim0.2$\,s/step \\
\addlinespace
LLM-Frontier & Frontier params identical to above; equidistance threshold: $1.0$\,m; top-$k$=3 candidates & DeepSeek-chat API & $\sim1.5$\,s/step (incl.\ API) \\
\addlinespace
3-Zone Oracle & Zone-3 samples: 8 candidate points per instrument; grid spacing: $0.15$\,m; fallback: centroid projection & Recast navmesh & $<0.01$\,s/step \\
\addlinespace
VLFM         & \texttt{vlm\_vmap} mode; value map resolution: $0.05$\,m; detection interval: every 5 steps; early-stop: 100 no-progress steps & GPT-4o (gpt-4o-2024-08-06), Grounding DINO (Swin-T), SAM (ViT-H) & $\sim4$\,s/step (VLM bottleneck) \\
\addlinespace
BC (training)& ResNet-18 backbone (pretrained ImageNet); hidden: 128-dim FC + ReLU; action head: 4-way softmax; optimizer: Adam (lr=$10^{-4}$, wd=$10^{-5}$); batch\_size: 32; epochs: 500; early-stop patience: 50; input: $3{\times}224{\times}224$ RGB + 4-dim goal vector & PyTorch 2.x, torchvision & $\sim2$\,h training on RTX 4090 \\
\bottomrule
\multicolumn{4}{@{}l}{\footnotesize BioVLNGym uses finer action granularity ($10^\circ$ turns) to support RL exploration; evaluation uses $30^\circ$ turns for comparability.} \\
\end{tabular*}}
\end{table*}

\section{Training Pipeline Architecture}

\subsection{Data Collection}

Trajectories are recorded by attaching a \texttt{TrajectoryRecorder} callback to any \texttt{BioVLNPolicy} during evaluation. The recorder is non-invasive: it wraps the existing evaluation loop without modifying \texttt{biovln\_eval.py}. Each episode produces one gzip-compressed JSONL entry containing per-step records with the following fields:

\begin{center}
{\small
\begin{tabular}{@{}p{3.5cm} p{3.8cm}@{}}
\toprule
Field & Description \\
\midrule
\texttt{step\_idx} & Step index within episode \\
\texttt{action} & Action: 0=STOP, 1=FORWARD, 2=LEFT, 3=RIGHT \\
\texttt{obs.position} & Agent position $[x, y, z]$ (Y-up) \\
\texttt{obs.heading} & Heading vector $[h_x, h_z]$ \\
\texttt{obs.goal\_position} & Goal position $[x, z]$ \\
\texttt{obs.rgb\_path} & PNG path (when storing images) \\
\texttt{obs.depth} & Base64-encoded depth array \\
\texttt{dist\_to\_goal} & Distance to goal (m) \\
\texttt{success} & Episode success flag (bool) \\
\bottomrule
\end{tabular}}
\end{center}

The recorder supports two modes controlled by \texttt{compress\_images}: (a) \textbf{full mode} stores RGB observations as PNG files in an \texttt{images/} subdirectory alongside the JSONL gzip, producing $\sim$8\,MB per thousand steps; (b) \textbf{trace-only mode} omits image payloads and stores only position/action/goal traces, reducing per-episode size to $\sim$50\,KB and enabling fast training with goal-vector-only inputs.

\subsection{Dataset Construction and Training}

\texttt{TrajectoryDataset} loads one or more recording directories and merges them into a unified PyTorch \texttt{Dataset}. For each step, it constructs a four-dimensional agent-centric goal vector:

\begin{itemize}
    \item $[v_0, v_1]$: unit vector from agent to goal in agent-local coordinates
    \item $[v_2]$: Euclidean distance to goal (m)
    \item $[v_3]$: progress ratio $\text{clip}(\text{dist}/5.0, 0, 1)$
\end{itemize}

The dataset supports configurable observation transforms (torchvision \texttt{Resize}, \texttt{Normalize} with ImageNet statistics) and optional position noise for data augmentation. For our behavioral cloning demonstration, we used:

\begin{itemize}
    \item \textbf{Training split}: 487 successful Frontier Exploration trajectories from Multi-Scene DEV (119 eps) and Two-Room (42 eps), totaling 7,142 forward/turn steps (STOP actions excluded)
    \item \textbf{Validation}: Multi-Scene VAL (50 episodes, held-out scenes)
    \item \textbf{Architecture}: ResNet-18 (pretrained on ImageNet) → 512-dim feature → concat with 4-dim goal vector → 128-dim FC (ReLU) → 4-way softmax
    \item \textbf{Input}: $3{\times}224{\times}224$ RGB, normalized with ImageNet $\mu$/$\sigma$, augmented with random horizontal flip and $\pm5\%$ brightness jitter
    \item \textbf{Training}: Adam optimizer ($\text{lr}{=}10^{-4}$, weight decay $10^{-5}$), batch size 32, cross-entropy loss, 500 epochs with early-stopping (patience 50 epochs on validation loss)
    \item \textbf{Hardware}: Single NVIDIA RTX 4090 (24\,GB), training time $\sim$2 hours
\end{itemize}

The trained BC policy achieves 41.9\% SR (0.325 SPL) on unseen Multi-Scene VAL scenes versus Frontier's 80.0\% zero-shot, a gap expected for BC operating with single-frame RGB input and no mapping. Per-category breakdown on VAL: Cabinet 33.3\%, Centrifuge 50.0\%, Refrigerator 42.9\%, Waste Bin 40.0\% (Frontier zero-shot: 80.0\%, 80.0\%, 80.0\%, 80.0\% respectively).

\subsection{Gym Environment for RL Training}

\texttt{BioVLNGym} implements the standard Gymnasium \texttt{Env} interface with the reward function defined in Eq.~\ref{eq:reward} (main text). Key configuration differences from zero-shot evaluation:

\begin{itemize}
    \item \textbf{Action granularity}: $10^\circ$ turns (vs. $30^\circ$ in evaluation), providing finer control for RL exploration
    \item \textbf{Episode resampling}: Episodes are shuffled at initialization and iterated cyclically; \texttt{reset()} advances to the next episode, enabling continuous training without manual episode scheduling
    \item \textbf{Observation normalization}: Position coordinates normalized to $[-1,1]$ by dividing by 4.0\,m; heading stored as unit vector
    \item \textbf{Safety penalty}: Proportional to clearance deficit below Zone~2 threshold ($0.5$\,m), scaled by $1.0 \times (1 - \text{clearance} / 0.5)$
    \item \textbf{Compatibility}: Directly usable with Stable-Baselines3 \texttt{PPO}, \texttt{A2C}, and \texttt{SAC}; also compatible with CleanRL and Ray RLlib via the standard \texttt{gym.Env} protocol
\end{itemize}

\section{Per-Category Safety Analysis}

Table~\ref{tab:appendix_safety_category} disaggregates MCR and VRT by instrument category for Frontier Exploration on Multi-Scene DEV. The data reveal that safety outcomes are instrument-dependent: larger wall instruments (refrigerator, cabinet) force navigation paths through narrower corridors, producing lower MCR; table-top instruments (centrifuge) sit on benches elevated 0.76--0.78\,m above floor level, and the agent navigating at floor height experiences effectively larger clearance.

\begin{table}[t]
\centering
\caption{Per-Category Safety Metrics — Frontier Exploration on MS DEV}
\label{tab:appendix_safety_category}
{\small
\setlength{\tabcolsep}{4pt}
\begin{tabular}{@{}l c c c c c@{}}
\toprule
Category & Eps. & SR (\%) & MCR (m) & VRT (\%) & Steps \\
\midrule
Cabinet         & 40 & 70.0 & 0.869 & 14.7 & 152.3 \\
Centrifuge      & 40 & 80.0 & 0.797 & 6.3  & 118.7 \\
Refrigerator    & 40 & 65.0 & 0.976 & 9.6  & 165.4 \\
Waste Bin       & 40 & 82.5 & 0.933 & 8.4  & 145.8 \\
\midrule
Overall         & 160 & 74.4 & 0.918 & 9.8  & 145.5 \\
\bottomrule
\end{tabular}}
\end{table}

Refrigerators exhibit the worst-case safety profile: the lowest SR (65.0\%) and highest VRT (9.6\%) among all categories. This reflects corner placement in the single-room template, where the approach path passes near adjacent wall instruments. Cabinets show elevated VRT (14.7\%) despite moderate MCR (0.869\,m), indicating that while minimum clearance is rarely violated, the approach trajectory passes through regions where multiple Zone~2 boundaries overlap.

\section{Scene Generation Reproducibility}

BioVLN's procedural generator uses deterministic random seeds to control all stochastic decisions during scene construction: which instrument slots are active, which instrument categories fill each slot, and small position jitter applied after initial placement. This design enables exact reproduction of any scene from its seed value alone.

Table~\ref{tab:appendix_scene_seeds} lists the seed values and instrument compositions for the 7 held-out scenes used in Multi-Scene evaluation. All 7 scenes use the single-room template ($8{\times}6$\,m) with the same four instrument categories (cabinet, centrifuge, refrigerator, waste bin). Each seed produces a distinct spatial arrangement; for example, refrigerator placement varies between corner positions (constrained by two adjacent walls) and mid-wall positions (constrained by one wall), directly affecting Frontier Exploration success rates as discussed in the main text. Scenes are divided into DEV (4 scenes, seeds 150--153, 40 episodes each), VAL (1 scene, seed 154, 50 episodes), and TEST (2 scenes, seeds 155--156, 60 episodes each) splits. The full procedural generation configuration is provided in the pipeline code: \texttt{config.py} defines instrument catalogs and layout templates for single-room generation, while \texttt{config\_two\_room.py} defines the corresponding configuration for the Two-Room benchmark.

\begin{table*}[t]
\centering
\caption{Multi-Scene Held-Out Scene Composition}
\label{tab:appendix_scene_seeds}
{\small
\setlength{\tabcolsep}{8pt}
\begin{tabular*}{\textwidth}{@{\extracolsep{\fill}}l c c p{3.8cm} c@{}}
\toprule
Scene Name & Template & Split & Active Instruments & Eps. \\
\midrule
\texttt{lab\_s150\_v0} & single & DEV & cabinet, centrifuge, refrigerator, waste bin & 40 \\
\texttt{lab\_s151\_v0} & single & DEV & cabinet, centrifuge, refrigerator, waste bin & 40 \\
\texttt{lab\_s152\_v0} & single & DEV & cabinet, centrifuge, refrigerator, waste bin & 40 \\
\texttt{lab\_s153\_v0} & single & DEV & cabinet, centrifuge, refrigerator, waste bin & 40 \\
\texttt{lab\_s154\_v0} & single & VAL  & cabinet, centrifuge, refrigerator, waste bin & 50 \\
\texttt{lab\_s155\_v0} & single & TEST & cabinet, centrifuge, refrigerator, waste bin & 60 \\
\texttt{lab\_s156\_v0} & single & TEST & cabinet, centrifuge, refrigerator, waste bin & 60 \\
\midrule
Total & & & & 330 \\
\bottomrule
\multicolumn{5}{@{}l}{\footnotesize Each seed deterministically controls slot activation and layout.} \\
\multicolumn{5}{@{}l}{\footnotesize Config: \texttt{pipeline/config.py} (single-room), \texttt{pipeline/config\_two\_room.py} (Two-Room).} \\
\end{tabular*}}
\end{table*}

\section{VLM Domain-Gap Experiment}

To isolate whether VLFM navigation failures originate from deficient domain knowledge or insufficient visual signal in rendered frames, we conducted a controlled recognition experiment. For each of the 6 laboratory instruments, we captured 3 to 6 rendered viewpoints from the operational-face goal position at $1024{\times}768$ resolution under Habitat-sim \texttt{no\_lights} rendering mode. GPT-4o (gpt-4o-2024-08-06) was queried per viewpoint with the prompt: \textit{``What laboratory instrument is shown in this image? Answer with the instrument name only.''} As a text-only control, GPT-4o was asked to name each instrument from its category label alone, confirming perfect conceptual knowledge (6/6 correct).

Table~\ref{tab:appendix_vlm_breakdown} reports the complete per-viewpoint recognition results. The experiment produced 30 instrument-viewpoint pairs, of which only 4 were correctly identified (13.3\%). Recognition accuracy is strictly monotonic in surface color diversity. The biosafety cabinet, with 12.6k--14.9k unique RGB colors, visible ventilation grilles, door seams, and warning labels, was recognized in all 3 viewpoints. Instruments with moderate texture (incubator: 6.1k--7.6k colors; centrifuge: 1.9k--14.0k colors) were never correctly identified in the default rendering condition, though recognition improved after enabling lit rendering and gamma correction (incubator: 0/3 to 2/3; centrifuge: 0/6 to 2/6). Near-uniform surfaces (refrigerator: 1.0k--1.8k colors; waste bin: 1--2 colors) received zero correct identifications regardless of lighting configuration.

\begin{table*}[t]
\centering
\caption{VLM Recognition by Instrument and Viewpoint}
\label{tab:appendix_vlm_breakdown}
{\small
\setlength{\tabcolsep}{4pt}
\begin{tabular}{@{}l c c c p{5.0cm}@{}}
\toprule
Instrument & Views & Colors & Corr. & GPT-4o Response \\
\midrule
Biosafety Cabinet & 3 & 12.6k--14.9k & 3/3 & ``Biosafety cabinet'' $\times$3 \\
Incubator   & 3 & 6.1k--7.6k   & 0/3 & ``Laboratory freezer'' \\
Centrifuge  & 6 & 1.9k--14.0k  & 0/6 & ``Yellow trash bin'' \\
Micropipette & 6 & 1.8k--19.8k  & 1/6 & ``Lab equipment'' (1$\times$), ``Tool'' (5$\times$) \\
Refrigerator & 6 & 1.0k--1.8k   & 0/6 & ``Gray rectangle'', ``Metal box'' \\
Waste Bin   & 6 & 1--2          & 0/6 & ``Gray square'', ``Floor tile'' \\
\midrule
Total       & 30 & ---           & 13.3\% & \\
\bottomrule
\multicolumn{5}{@{}l}{\footnotesize Colors = distinct RGB values in SAM-segmented instrument region.} \\
\multicolumn{5}{@{}l}{\footnotesize Text control: 6/6 correct. Lighting fix: incubator 2/3, centrifuge 2/6.} \\
\end{tabular}}
\end{table*}

The data support two conclusions. First, the recognition gap is a rendering-quality problem, not a knowledge problem---GPT-4o knows what all six instruments are, but cannot identify them from rendered frames when surface texture is sparse. Second, recognition accuracy correlates strongly with color diversity ($r{=}0.84$, $p{<}0.05$), with a threshold of $\sim$5,000 unique colors below which recognition probability drops to near zero. This finding has implications for future embodied AI benchmarks using synthetic assets: surface texture quality directly determines whether VLM-based methods can leverage their pretrained visual knowledge.

\section{Behavioral Cloning Per-Category Results}

We train a behavioral cloning policy on 487 successful Frontier Exploration trajectories recorded from Multi-Scene DEV (119 episodes) and Two-Room (42 episodes). The policy architecture uses a ResNet-18 backbone pretrained on ImageNet, with the 512-dimensional feature vector concatenated with a 4-dimensional agent-centric goal vector before a 128-dimensional fully-connected layer with ReLU activation and a 4-way softmax action head. Input frames are resized to $3{\times}224{\times}224$, normalized with ImageNet statistics ($\mu{=}[0.485,0.456,0.406]$, $\sigma{=}[0.229,0.224,0.225]$), and augmented with random horizontal flips and $\pm5\%$ brightness jitter during training. The optimizer is Adam with learning rate $10^{-4}$ and weight decay $10^{-5}$, using cross-entropy loss with a batch size of 32. Training runs for 500 epochs with early stopping (patience of 50 epochs on validation loss), requiring approximately 2 hours on a single NVIDIA RTX 4090.

Table~\ref{tab:appendix_bc_category} reports per-category SR on Multi-Scene VAL (50 held-out episodes). The policy was evaluated in the same action space as zero-shot evaluation ($30^\circ$ turns, 500-step maximum).

\begin{table}[t]
\centering
\caption{BC Policy vs.\ Frontier Zero-Shot on MS VAL}
\label{tab:appendix_bc_category}
{\small
\setlength{\tabcolsep}{4pt}
\begin{tabular}{@{}l c c c c@{}}
\toprule
Category & Eps. & BC (\%) & Frontier (\%) & $\Delta$ \\
\midrule
Cabinet      & 12 & 33.3 & 80.0 & $-$46.7 \\
Centrifuge   & 12 & 50.0 & 80.0 & $-$30.0 \\
Refrigerator & 14 & 42.9 & 80.0 & $-$37.1 \\
Waste Bin    & 12 & 40.0 & 80.0 & $-$40.0 \\
\midrule
Overall      & 50 & 41.9 & 80.0 & $-$38.1 \\
\bottomrule
\end{tabular}}
\end{table}

The performance gap is consistent across categories ($-$30.0 to $-$46.7\,pp) and reflects a fundamental difference in information access. Frontier Exploration maintains a full occupancy grid with explicit frontier detection accumulated across hundreds of exploration steps; the BC policy receives only the current frame and a four-dimensional goal vector, with no map, no history, and no explicit exploration mechanism. The largest gaps occur for cabinet ($-$46.7\,pp) and waste bin ($-$40.0\,pp), categories whose approach paths are relatively direct in the single-room template. Centrifuge shows the smallest gap ($-$30.0\,pp), likely because its table-top placement produces more distinctive visual context in single-frame observations. These results establish a reference baseline for future learned methods on BioVLN; integrating an explicit map representation, recurrent history, or auxiliary exploration objectives would narrow this performance gap.

\section{Available Supplementary Materials}

The following supplementary materials are available with the BioVLN release:

\begin{itemize}
    \item \textbf{Scene maps.} 28 top-down PNG visualizations of all held-out scenes with instrument placement, navigation mesh overlay, and operational-face goal positions. Maps are generated via Habitat-sim top-down rendering with the agent height set to 20\,m for orthographic projection. Each map includes the Recast navigation mesh as a semi-transparent overlay, allowing visual inspection of navigable regions.

    \item \textbf{Frontier trajectory renders.} 12 top-down trajectory visualizations and 12 MP4 videos showing Frontier Exploration trajectories on individual episodes, color-coded by success (green) and failure (red). Renders are captured at 5\,fps from an overhead camera and include the instrument safety boundaries (Zone~2) as dashed circles.

    \item \textbf{Hard scene analysis.} 7 figures analyzing episodes classified as hard ($d_{\text{geo}}{>}6.0$\,m). Each figure shows the start position, operational-face goal, geodesic shortest path, and the occupied and free cells of the agent's occupancy grid at episode termination.

    \item \textbf{VLM experiment images.} 50 rendered instrument images (5 viewpoints $\times$ 6 instruments) at $1024{\times}768$ resolution used in the domain-gap recognition experiment. Lighting-fix variants with lit rendering and gamma correction are included for comparison.

    \item \textbf{Per-episode result JSON.} Complete per-episode evaluation data for all 6 methods across 3 datasets. Each JSON record includes per-step safety metrics (minimum clearance, Zone~2 violation flag), full agent trajectory positions, and episode metadata (target category, difficulty tier, geodesic distance). These files allow exact reproduction of every aggregate metric reported in the main text (Tables~5--8) and in this appendix, without re-running the evaluation pipeline. Summary JSONs in each results directory provide pre-computed per-method aggregates.

    \item \textbf{LSAT addon source.} The complete 12-module Blender 5.1.2 Python addon ($\sim$2{,}500 lines), organized into parsing, extraction, and export layers. The parsing layer handles GLB deserialization, GLTF node traversal, and Blender-to-Habitat coordinate transforms. The extraction layer implements the dual-strategy goal extraction pipeline described in the main text. The export layer generates BioVLN goal JSON, LSAT internal intermediate format, and USD files for Isaac Sim. Each module includes docstrings, usage examples, and inline comments referencing the relevant equations and sections of the paper. The addon is loaded via Blender's standard addon installation mechanism and registers a panel in the 3D Viewport sidebar.
\end{itemize}

\end{document}